\documentclass[conference]{IEEEtran}
\IEEEoverridecommandlockouts
\usepackage{multirow}
\usepackage{cite}
\usepackage{amsmath,amssymb,amsfonts}
\usepackage{algorithmic}
\usepackage{graphicx}
\usepackage{booktabs}
\usepackage{textcomp}
\usepackage{array}
\newcolumntype{L}[1]{>{\raggedright\arraybackslash}p{#1}}
\usepackage{xcolor}
\usepackage[colorlinks=true,linkcolor=blue,citecolor=black,urlcolor=blue]{hyperref}

\def\BibTeX{{\rm B\kern-.05em{\sc i\kern-.025em b}\kern-.08em
    T\kern-.1667em\lower.7ex\hbox{E}\kern-.125emX}}
\begin{document}

\title{Edge-Efficient Two-Stream Multimodal Architecture for Non-Intrusive Bathroom Fall Detection}


\author{
\IEEEauthorblockN{
Haitian Wang\textsuperscript{*}\thanks{\textsuperscript{*} Corresponding author},
Yiren Wang,
Xinyu Wang,
Sheldon Fung,
Atif Mansoor
}
\IEEEauthorblockA{
The University of Western Australia, 35 Stirling Highway, Crawley, WA 6009, Australia\\
\{haitian.wang, bohr.wang, xinyu.wang, sheldon.feng, atif.mansoor\}@uwa.edu.au\\
}
}


\maketitle

\begin{abstract}
Falls in wet bathroom environments are a major safety risk for seniors living alone. Recent work has shown that mmWave-only, vibration-only, and existing multimodal schemes, such as vibration-triggered radar activation, early feature concatenation, and decision-level score fusion, can support privacy-preserving, non-intrusive fall detection. However, these designs still treat motion and impact as loosely coupled streams, depending on coarse temporal alignment and amplitude thresholds, and do not explicitly encode the causal link between radar-observed collapse and floor impact or address timing drift, object drop confounders, and latency and energy constraints on low-power edge devices. To this end, we propose a two-stream architecture that encodes radar signals with a Motion--Mamba branch for long-range motion patterns and processes floor vibration with an Impact--Griffin branch that emphasizes impact transients and cross-axis coupling. Cross-conditioned fusion uses low-rank bilinear interaction and a Switch--MoE head to align motion and impact tokens and suppress object-drop confounders. The model keeps inference cost suitable for real-time execution on a Raspberry Pi 4B gateway. We construct a bathroom fall detection benchmark dataset with frame-level annotations, comprising more than 3~h of synchronized mmWave radar and triaxial vibration recordings across eight scenarios under running water, together with subject-independent training, validation, and test splits. On the test split, our model attains 96.1\% accuracy, 94.8\% precision, 88.0\% recall, a 91.1\% macro F1 score, and an AUC of 0.968. Compared with the strongest baseline, it improves accuracy by 2.0 percentage points and fall recall by 1.3 percentage points, while reducing latency from 35.9\,ms to 15.8\,ms and lowering energy per 2.56\,s window from 14200\,mJ to 10750\,mJ on the Raspberry Pi 4B gateway.
\end{abstract}


\vspace{-2mm}
\section{Introduction}
\label{sec:intro}

The proportion of seniors living alone continues to rise, making home safety an increasingly critical concern~\cite{leong2018world, strini2021fall}. Bathrooms represent the highest-risk environment due to wet floors, hard surfaces, and confined layouts, with studies reporting that the majority of household falls occur in this setting~\cite{wang2020elderly}. Commercial solutions have not addressed this challenge effectively: caregiver monitoring is costly and lacks scalability~\cite{cardoso2020care}, wearable devices are often neglected or cannot function reliably in wet conditions\cite{wang2025p2mfds}, and camera-based or microphone-based systems introduce privacy concerns that prevent their use in bathrooms~\cite{kittiyanpunya2023mmwave}. These limitations have directed attention toward non-intrusive, privacy-preserving sensing approaches that can operate reliably in such environments~\cite{shen2025falldetection,gorce2025review, Pasa10869610}.

Among non-intrusive sensing technologies, radar and floor-mounted vibration sensors have emerged as two of the most studied options for unobtrusive fall detection in private indoor environments ~\cite{mubashir2013survey,igual2013challenges,erol2017rangedoppler,clemente2020seismic}. Continuous-wave and millimeter-wave radars provide range--Doppler and micro-motion signatures that capture pre-impact trajectories and body-posture evolution, yet existing systems still report false alarms when objects are dropped and when strong specular reflections or multi-path clutter distort the Doppler trace in confined rooms ~\cite{erol2017rangedoppler,shrestha2019radaractivities,sadreazami2020standoff}. Floor vibration approaches exploit structural coupling to obtain a high signal-to-noise ratio at impact and can detect events over a wide area with a small number of sensors ~\cite{zigel2009fall,litvak2008floor,shao2020floor,clemente2020seismic}. However, they observe only the impact phase; their response is highly dependent on floor construction and sensor mounting, and non-human impacts with similar energy profiles remain difficult to reject ~\cite{zigel2009fall,litvak2008floor,shao2020floor,clemente2020seismic}. Multimodal designs that combine inertial or visual streams with other sensors often use simple early feature concatenation or decision-level score averaging, which treats each sensor as an independent classifier and does not explicitly model the causal relationship between a motion pattern and the subsequent impact or correct for sampling-rate mismatch and clock drift ~\cite{abbate2012smartphone,yadav2023multimodal,wang2023egofalls,wang2024multispectral,li2018collaborative}. Most of these systems are evaluated under laboratory conditions and only a few report end-to-end latency or energy on embedded processors, so their suitability for continuous bathroom monitoring on low-power edge platforms remains unclear.

\begin{figure*}[!t]
    \centering
    \scalebox{1}[0.85]{%
    \includegraphics[width=0.90\linewidth]{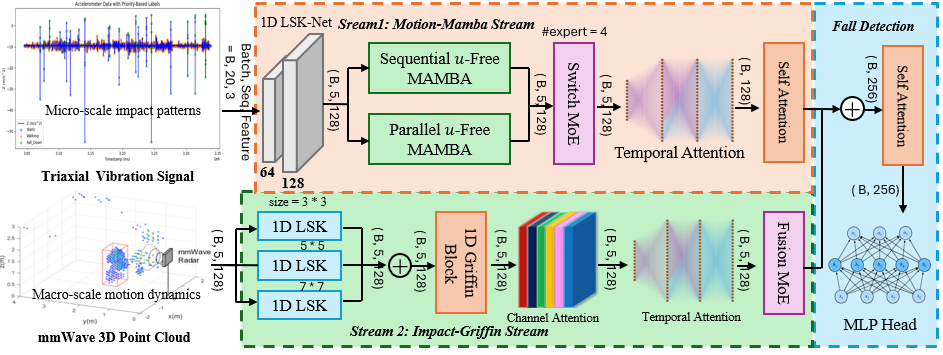}%
    }
    \vspace{-4mm}
    \caption{Overall architecture of the proposed fall detection framework. Radar signals are processed by the Motion--Mamba stream and vibration signals by the Impact--Griffin stream. The two branches are fused through cross–conditioned interaction, low–rank bilinear coupling, and a fusion Switch--MoE.}
    \label{fig:method_overview}
    \vspace{-7mm}
\end{figure*}

To address these gaps, we propose a two-stream mmWave radar and floor vibration architecture whose fusion is explicitly alignment-aware and temporally coherent while remaining suitable for low-power embedded deployment in wet, cluttered bathrooms. The radar stream uses large, selective kernels and a linear-time state-space module to capture pre-impact motion and collapse over several seconds, and to suppress clutter patterns produced by specular reflections and water. The vibration stream applies selective filtering, gated recurrent memory, and inter-channel weighting to preserve sharp impact transients and short-duration ringing while absorbing axis-dependent coupling into the feature representation. Motion and impact tokens are then cross-conditioned in time and combined by a low-rank bilinear module with a compact Switch--MoE adapter. This links radar-observed collapse to the subsequent impact, reduces false alarms from object drops, improves robustness to sampling mismatch and packet loss, and keeps fusion cost proportional to sequence length. Since there is no public benchmark that jointly records mmWave radar and floor vibration in bathrooms, we also construct a benchmark dataset with more than three hours of synchronized recordings that cover eight representative fall and non-fall scenarios under running water, together with frame-level labels and subject-independent train, validation, and test splits. On the held-out test split, the proposed method reaches 96.1\% window-level accuracy, 94.8\% precision, 88.0\% recall, and a 91.1\% macro F1 score with an AUC of 0.968. Compared with the strongest baseline on this dataset, it raises accuracy from 94.1\% to 96.1\%, increases fall recall from 86.7\% to 88.0\%, reduces latency from 35.9\,ms to 15.8\,ms, and lowers the energy per 2.56\,s window from 14200\,mJ to 10750\,mJ on the Raspberry Pi~4B gateway.

Our main contributions are as follows:
\begin{itemize}
    \item We propose and implement an end-to-end sensing and inference pipeline for privacy-preserving bathroom elderly fall detection on low-power edge hardware.
    \item We design a two-stream mmWave radar and floor-vibration network architecture that performs alignment-aware multimodal fusion for robust fall detection.
\end{itemize}


\vspace{-2mm}
\section{Methodology}

Our method converts raw radar and vibration signals into synchronized feature sequences, processes them through two modality–specific streams, and fuses the resulting tokens for fall detection, as illustrated in Fig.~\ref{fig:method_overview}.

\vspace{-2mm}
\subsection{Multimodal Signal Preprocessing}
\label{sec:preprocess}

Raw triaxial vibration $x_v[n]\in\mathbb{R}^{3}$ and mmWave radar streams $x_r(t)$ are first mapped to a common rate $f_s=100$~Hz. Radar frames are converted to per-frame kinematic descriptors and linearly interpolated to a uniform sequence $x_r^{\text{uni}}[n]$. The inter sensor delay $\Delta\tau$ is estimated from vibration and radar energy envelopes $e_v[n]$ and $e_r[n]$ using GCC-PHAT~\cite{knapp1976gccphat}, 
\vspace{-2mm}
\begin{equation}
\hat{\Delta\tau}
=\arg\max_{\tau}\,
\mathcal{F}^{-1}\!\left\{
\frac{X_v(\omega)\,X_r^{*}(\omega)}
{|X_v(\omega)\,X_r(\omega)|}
\right\}[\tau],
\end{equation}
with $X_v$ and $X_r$ the discrete time Fourier transforms of $e_v$ and $e_r$. This estimator is less sensitive to tiled-bathroom reverberation than direct time-domain correlation. Residual clock drift is modeled by a linear function $\tau(n)=\alpha n+\beta$ fitted to local correlation peaks, which gives a simple affine warp of $x_r^{\text{uni}}[n]$ instead of a full dynamic time warping step~\cite{sakoe1978dtw}.

Aligned vibration is denoised by discrete wavelet shrinkage and a short Hampel filter, then converted to a scalar envelope $e_v[n]=\sqrt{\sum_{c}(x_v^{(c)}[n])^{2}}$ used for supervision and centering. Radar frames are background-subtracted in the range–Doppler map, thresholded by an adaptive noise floor, clustered in $(r,\dot r)$, and summarized as a compact kinematic vector $\phi_t=[\bar r_t,\overline{|\dot r|}_t,\sigma_{r,t},\sigma_{\dot r,t},E_t]$, from which we also derive a smooth radar envelope $e_r[n]$. A 1~s sliding correlation between $e_v[n]$ and $e_r[n]$ defines a consistency weight $\gamma[n]$ that down-weights windows with poor cross-modal agreement during training. All feature channels are robustly normalized, and signals are sliced into windows of $T_w=2.56$~s with $50\%$ overlap ($L=256$ samples at $f_s=100$~Hz), shifting fall windows so that the peak of $e_v[n]$ lies near the center of the LSK1D receptive field. We apply light-label-preserving augmentation (small Gaussian jitter, mild time stretching, and narrow-band dropout around the dominant vibration resonance) to improve robustness to mounting conditions and subject variability while preserving the causal relationship between radar motion and floor impact.

\vspace{-2mm}
\subsection{Stream~1: Motion--Mamba Radar Branch}
\label{sec:motion_mamba}

Given the preprocessed radar sequence $\mathbf{X}_r\in\mathbb{R}^{T\times C_0}$, the Motion--Mamba branch first applies two LSK1D blocks that map it to $\mathbf{Y}\in\mathbb{R}^{T'\times C}$. Each block uses three depthwise 1D convolution branches with kernel sizes $3,7,11$ and dilations $1,2,3$, followed by a shared pointwise projection. Channelwise global average pooling with a small gating MLP produces soft weights over branches so that the network can emphasize short transients or slower approach motion. The second block uses temporal stride $2$ at $f_s=100$~Hz, which reduces sequence length and provides a receptive field of several hundred milliseconds with low computation.

Long range temporal structure in $\mathbf{Y}$ is modeled by two stacked Mamba2Block1D modules~\cite{gu2023mamba}. Each block combines a causal depthwise convolution with kernel size $5$ and a gated state space recurrence with state dimension $d_{\text{state}}=64$ and expansion factor $2$. At time index $t$, with input $\mathbf{x}_t$ and hidden state $\mathbf{h}_t$, the recurrent core is
\vspace{-3mm}
\begin{equation}
\mathbf{h}_t = \mathbf{g}_t \odot (\mathbf{A}\mathbf{h}_{t-1}) + (1-\mathbf{g}_t)\odot(\mathbf{B}\mathbf{x}_t),\quad
\mathbf{y}_t = \mathbf{C}\mathbf{h}_t + \mathbf{D}\mathbf{x}_t,
\end{equation}
where $\mathbf{g}_t$ is an input dependent gate and $\odot$ denotes elementwise product. The scan implementation evaluates this recurrence in linear time with constant memory per token so that pre impact motion, collapse, and short post impact motion are captured over several seconds without the cost of full attention layers.

On top of the state-space features we use a Switch--MoE temporal adapter with attention pooling to form a motion token $\mathbf{m}\in\mathbb{R}^{C}$. A single head self attention over a window of seven frames refines local dynamics and a router with $E{=}4$ expert feed forward networks selects one expert per time step with a load balancing loss, so that experts specialize to different motion regimes while the radar branch remains efficient for deployment in wet bathroom environments.


\vspace{-2mm}
\subsection{Stream~2: Impact--Griffin Vibration Branch}
\label{sec:impact_griffin}

This branch maps the preprocessed triaxial vibration sequence $\mathbf{X}_v\in\mathbb{R}^{T\times 3}$ to an impact aware token. A two stage LSK1D front end lifts $\mathbf{X}_v$ to $\mathbf{V}\in\mathbb{R}^{T'\times C}$ using three depthwise branches with $(k_m,d_m)\in\{(3,1),(5,2),(7,3)\}$ and content dependent weights from channelwise global averages. The short path focuses on the impact onset and the wider paths capture the first ringing cycles, while the selector suppresses quasi stationary structural tones that a fixed kernel would pass. Stride $2$ at $f_s=100$ Hz reduces sequence length while keeping the main impact within a few frames so the cost stays within the budget of an embedded node.

On $\mathbf{V}$ we apply a 1D Griffin block that first mixes local context and then updates a gated long term memory. A depthwise convolution over $w=9$ frames followed by a pointwise projection produces $\mathbf{v}_t$. The GLRU cell uses
\begin{small}
\vspace{-2mm}
\begin{equation}
\vspace{-2mm}
\label{eq:griffin_update_compact}
\begin{aligned}
\tilde{\mathbf{h}}_t &=
\phi\!\big(\mathbf{W}_h\mathbf{v}_t
+\mathbf{U}_h(\mathbf{g}_t\odot\mathbf{h}_{t-1})\big),
\mathbf{h}_t &=
(1-\boldsymbol{\alpha}_t)\odot\mathbf{h}_{t-1}
\end{aligned}
\end{equation}
\end{small}
where $\mathbf{g}_t=\sigma(\mathbf{W}_g\mathbf{v}_t+\mathbf{U}_g\mathbf{h}_{t-1})$ controls how much of $\mathbf{h}_{t-1}$ enters the candidate and $\boldsymbol{\alpha}_t=\sigma(\mathbf{W}_\alpha\mathbf{v}_t+\mathbf{b}_\alpha)$ sets the update rate. This gives slow response for low amplitude motion while allowing rapid changes at impact under a single hidden width. A residual projection with LayerNorm and a single head local self attention over $w_a=11$ frames refine $\mathbf{h}_t$ by reweighting only nearby samples, so modeling capacity is concentrated on impact neighborhoods rather than the full sequence.

To adapt to posture and mounting we apply inter channel attention based on a robust covariance of $\mathbf{H}\in\mathbb{R}^{T'\times C}$, the stack of $\mathbf{h}_t$. We estimate a shrunk covariance and low rank factorization
\vspace{-2mm}
\begin{small}
\vspace{-2mm}
\begin{equation}
\vspace{-2mm}
\label{eq:ica_weight}
\begin{aligned}
\mathbf{\Sigma}
&=\frac{1}{T'}\sum_{t=1}^{T'}(\mathbf{h}_t-\bar{\mathbf{h}})
(\mathbf{h}_t-\bar{\mathbf{h}})^{\top}
+\epsilon\mathbf{I}\approx\mathbf{P}\mathbf{P}^{\top},\\
\mathbf{s}
&=\sigma\big(\mathbf{W}_2\phi(\mathbf{W}_1\mathbf{d})\big),
\quad
\mathbf{d}=\operatorname{diag}(\mathbf{P}\mathbf{P}^{\top}),
\end{aligned}
\end{equation}
\end{small}
with $\mathbf{P}\in\mathbb{R}^{C\times r}$ and $r=8$. The vector $\mathbf{s}\in\mathbb{R}^{C}$ amplifies axes that co activate at impact in the whitened space and the refined sequence is $\widetilde{\mathbf{H}}=\mathbf{H}\odot\mathbf{s}^{\top}$. Compared with simple squeeze and excitation, this covariance normalized weighting is insensitive to per axis scale changes caused by distance or floor stiffness and focuses on coherent cross axis patterns.

We apply a single query temporal attention to $\widetilde{\mathbf{H}}$. A learned query $\mathbf{q}\in\mathbb{R}^{C}$ produces keys and values by linear projections, attention weights over $T'$ define a convex combination of frames, and the result is an impact token $\mathbf{i}\in\mathbb{R}^{C}$ followed by a final linear projection. This concentrates fall related energy into a fixed width descriptor that is directly consumed by the fusion module instead of averaging the whole window.


\begin{figure}[!b]
    \vspace{-6mm}
    \centering
    \scalebox{1}[0.7]{%
    \includegraphics[width=0.95\linewidth]{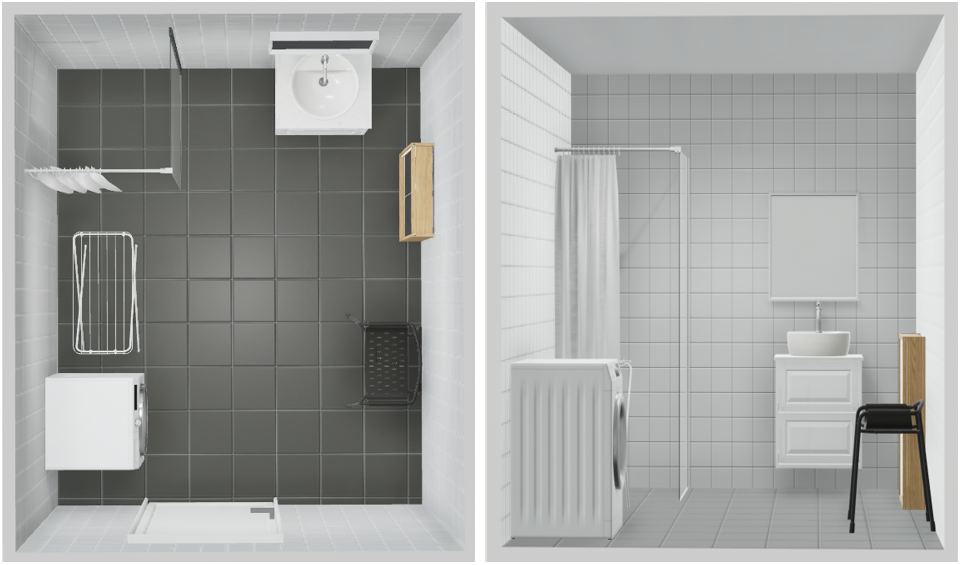}
    }
    \vspace{-4mm}
    \caption{Experimental bathroom mock-up ($3.70{\times}2.50~\mathrm{m}$). 
    Left: top-down floor plan showing the shower bay, fixtures, and reference dimensions. 
    Right: front view illustrating the sink, mirror, partition, and furniture placement. }
    \label{fig:floorplan_setup}
    \vspace{-4mm}
\end{figure}

\begin{figure}[!b]
    \centering
    \scalebox{1}[0.7]{%
    \includegraphics[width=0.95\linewidth]{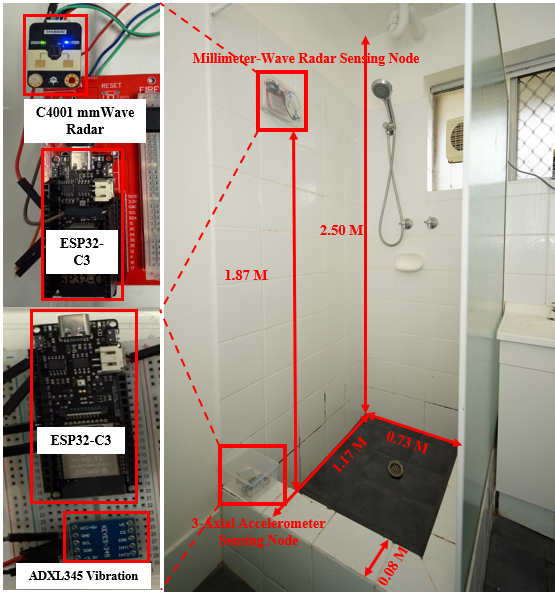}
    }
    \vspace{-4mm}
    \caption{Site photograph and hardware. Left: C4001 radar and ADXL345 node electronics (ESP32-C3 MCU). Right: installed locations—radar on the upper wall ($1.87~\mathrm{m}$ height), vibration node on the shower platform ($0.18~\mathrm{m}$ above floor); annotated room dimensions.}
    \label{fig:site_photo}
    \vspace{-8mm}
\end{figure}

\begin{table*}[!t]
\caption{Scenario-wise evaluation in the wet bathroom environment. For each scenario we report per-class precision and recall for Non--Fall and Fall windows and window-level Accuracy, Balanced Accuracy and Macro F1. The bottom block summarizes robustness tests of the full model under synthetic time misalignment and modality dropouts across all scenarios.}
\vspace{-2mm}
\centering
\setlength{\tabcolsep}{4.5pt}
\begin{tabular}{l c c cc cc c c c}
\toprule
\multirow{2}{*}{Scenario} & \multirow{2}{*}{Eval. Windows (N)} & \multicolumn{2}{c}{Non--Fall Metrics (\%)} & \multicolumn{2}{c}{Fall Metrics (\%)} & \multicolumn{3}{c}{Window-level (\%)}\\
\cmidrule(lr){3-4} \cmidrule(lr){5-6} \cmidrule(lr){7-9}
 &  & Precision & Recall & Precision & Recall & Acc. & Bal. Acc. & Macro F1\\
\midrule
\multicolumn{9}{l}{\emph{Per-scenario performance}}\\
\midrule
Empty Bathroom         & 218  & 97.00 & 92.50 & 96.20 & 88.10 & 97.10 & 90.30 & 93.33\\
Light Object Drop      & 498  & 94.80 & 93.00 & 92.00 & 87.60 & 96.38 & 90.30 & 91.80\\
Heavy Object Drop      & 380  & 90.10 & 77.50 & 85.00 & 64.40 & 91.30 & 70.95 & 78.30\\
Normal Walking         & 1188 & 97.20 & 95.60 & 96.10 & 91.50 & 97.83 & 93.55 & 95.08\\
Bent Posture Walk      & 896  & 94.00 & 88.90 & 91.10 & 78.80 & 94.93 & 83.85 & 87.94\\
Wall--Supported Walk   & 610  & 96.20 & 90.50 & 95.30 & 83.70 & 97.08 & 87.10 & 91.19\\
Static Standing        & 617  & 97.30 & 90.00 & 94.90 & 84.20 & 96.35 & 87.10 & 91.37\\
Squatting              & 291  & 93.70 & 97.60 & 97.00 & 95.90 & 96.27 & 96.75 & 96.03\\
\midrule
Macro mean across scenarios & --   & 95.04 & 90.70 & 93.45 & 84.28 & 95.91 & 87.49 & 90.63\\
\midrule
\multicolumn{9}{l}{\emph{Aggregates and robustness across all scenarios}}\\
\midrule
All scenarios (clean, window-weighted)          & 4698 & 95.10 & 89.70 & 94.40 & 86.10 & 96.10 & 87.90 & 91.10\\
All scenarios + $\pm150$\,ms time shift         & 4698 & 93.20 & 85.80 & 93.50 & 84.20 & 95.00 & 85.00 & 89.30\\
All scenarios + 30\% radar frame dropout        & 4698 & 94.20 & 87.10 & 94.40 & 86.50 & 95.60 & 86.80 & 90.40\\
All scenarios + 30\% vibration window dropout   & 4698 & 93.60 & 86.00 & 93.00 & 84.50 & 94.70 & 85.25 & 88.80\\
\bottomrule
\end{tabular}
\label{tab:results_all}
\vspace{-6mm}
\end{table*}

\begin{figure}[!t]
    \centering
    \scalebox{1}[0.7]{%
    \includegraphics[width=0.99\linewidth]{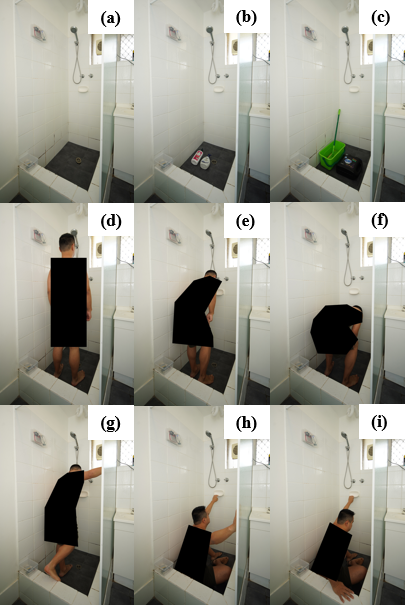}
    }
    \vspace{-4mm}
    \vspace{-3mm}
    \caption{Eight background scenarios and a representative intentional fall in the wet bathroom environment: (a) empty scene; (b) soap drop; (c) mop drop; (d) upright walking; (e) flexed‐torso walking; (f) wall‐assisted walking; (g) quiet standing; (h) squatting; (i) example of an intentional fall. All human activities were performed with the shower running.}
    \label{fig:Exp_behavior}
    \vspace{-8mm}
\end{figure}

\vspace{-2mm}
\subsection{Cross--Modal Fusion and Fall Decision}
\label{sec:fusion_detection}

Let $\widehat{\mathbf{Y}}\in\mathbb{R}^{T'\times C}$ and $\widetilde{\mathbf{H}}\in\mathbb{R}^{T'\times C}$ denote radar and vibration features. Fusion first applies local cross conditioning so that radar features are reinforced only when supported by nearby vibration activity and conversely. Conditioning radar on vibration uses a single head attention over a window of radius $w_c{=}6$, where
$\widehat{\mathbf{Y}}^{\star}_t
=\sum_{\tau=t-w_c}^{t+w_c}
\alpha_{t,\tau}\,\widetilde{\mathbf{H}}_{\tau}$ and the attention weights are
\begin{small}
\vspace{-2mm}
\begin{equation}
\vspace{-2mm}
\label{eq:xcond_icme}
\boldsymbol{\alpha}_t
=\mathrm{softmax}\!\left(
\frac{\widehat{\mathbf{Y}}_t\mathbf{W}_q^{(r)}(\widetilde{\mathbf{H}}\mathbf{W}_k^{(v)})^{\top}}{\sqrt{C}}
\right).
\end{equation}
\end{small}
The cross conditioned vibration sequence $\widetilde{\mathbf{H}}^{\star}$ is obtained by swapping roles. This promotes collapse patterns that are consistent across modalities and avoids the cost of global cross attention.

Cross conditioned sequences are summarized by single query temporal attention into a motion token $\mathbf{m}\in\mathbb{R}^{C}$ and an impact token $\mathbf{i}\in\mathbb{R}^{C}$. Their interaction is modeled by a low rank bilinear block that preserves multiplicative structure without the parameters of a full bilinear map. With projections $\mathbf{U},\mathbf{V}\in\mathbb{R}^{C\times d}$, $d{=}32$, and $K{=}4$ diagonal gates,
\begin{small}
\vspace{-2mm}
\begin{equation}
\vspace{-2mm}
\label{eq:mlb_icme}
\mathbf{z}
=\mathbf{W}_o\,\phi\!\left(
\sum_{k=1}^{K}\mathbf{D}_k\big(\mathbf{U}^{\top}\mathbf{m}\odot\mathbf{V}^{\top}\mathbf{i}\big)
\right),
\end{equation}
\end{small}
where $\phi$ is SiLU. The elementwise product activates only when both tokens carry energy in related subspaces which suppresses vibration only shocks and radar only clutter.

A compact Switch MoE then adapts the fusion rule. The router takes the concatenated vector $[\mathbf{m}\,\|\,\mathbf{i}\,\|\,\mathbf{z}]$, predicts routing probabilities, and selects one of $E{=}4$ expert MLPs for each window in a top one fashion. Each expert has hidden width $2C$ and a residual connection to $\mathbf{z}$. A linear head maps the adapted token to a logit and a sigmoid yields the fall probability. Training uses binary cross entropy together with a load balancing term for the router and an orthogonality penalty on $\mathbf{U}$ and $\mathbf{V}$, and all fusion operations remain $O(T'C)$.


\vspace{-2mm}
\section{Experiment}

This section evaluates the proposed framework in a controlled bathroom environment that replicates residential conditions with wet surfaces and confined geometry, emphasizing both robustness of radar–vibration fusion under realistic disturbances and practicality of deployment, including sensor placement, data collection, and a reproducible protocol.

\begin{figure}[!t]
    \centering
    \includegraphics[width=0.98\linewidth]{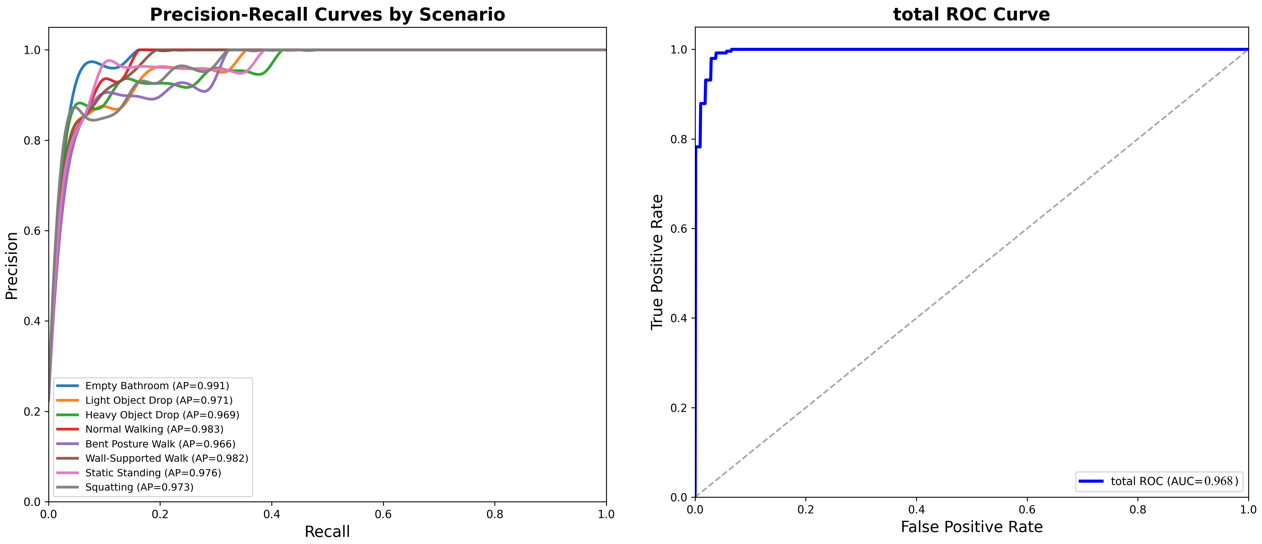}
    \vspace{-4mm}
    \caption{Scenario-wise precision--recall curves (left) and overall ROC curve (right). The micro-averaged ROC yields $\mathrm{AUC}=0.968$ which indicates high true-positive rate at low false-positive rate.}
    \label{fig:pr_roc}
    \vspace{-8mm}
\end{figure}

\vspace{-2mm}
\subsection{Experiments Environment and Sensor Placement}
\label{sec:Experiments_Environment}
\vspace{-1mm}

We conducted all trials in a full-scale bathroom mock-up ($3.70{\times}2.50~\mathrm{m}$) with ceramic tiles and a raised shower platform ($1.17{\times}0.73~\mathrm{m}$), as illustrated by the floor plan and front view in Fig.~\ref{fig:floorplan_setup}. Two sensing nodes were deployed (Fig.~\ref{fig:site_photo}): a triaxial vibration node (ADXL345 with ESP32-C3) rigidly fixed on the shower platform at floor level, and a mmWave node (C4001 radar with ESP32-C3) mounted on the wall at $1.87~\mathrm{m}$ to provide unobstructed coverage of the entire area, including the shower bay. Both nodes used sealed, waterproofed enclosures, shared ESP32-based time synchronization, and were calibrated in situ. During all trials the ESP32-C3 nodes performed sensing, timestamping and lightweight preprocessing, then transmitted buffered 2.56~s windows over Wi-Fi to a Raspberry~Pi~4B edge gateway, where the Motion--Mamba and Impact--Griffin streams and the fusion head were executed for on site inference and logging.

\begin{table*}[!t]
\caption{Latency and energy per 2.56$\,\mathrm{s}$ window for sensing nodes and inference back ends. The latency column lists preprocessing, model inference, and end-to-end detection delay in milliseconds. Power and energy are averaged over 200 windows in the wet bathroom environment.}
\vspace{-2mm}
\label{tab:sys_efficiency}
\centering

\scriptsize
\setlength{\tabcolsep}{2pt}
\renewcommand{\arraystretch}{1.1}

\resizebox{\textwidth}{!}{%
\begin{tabular}{l l l c c c c c c c}
\toprule
Subsystem & Hardware & Configuration & Params [M] & MACs [M] & Mod. size [MB] & Latency [ms] & $P_{\text{avg}}$ [W] & $E_{\text{win}}$ [mJ] & Cap. [win/s]\\
\midrule
\multirow{2}{*}{Sensing node}
& ESP32-C3 + ADXL345
& Vibration preprocessing
& --
& --
& --
& \begin{tabular}{@{}ccc@{}}\scriptsize pre & \scriptsize inf & \scriptsize end\\[-1pt]\scriptsize 0.9 & \scriptsize -- & \scriptsize 0.9\end{tabular}
& 0.32
& 820
& --\\[3pt]
& ESP32-C3 + C4001
& Radar front end
& --
& --
& --
& \begin{tabular}{@{}ccc@{}}\scriptsize pre & \scriptsize inf & \scriptsize end\\[-1pt]\scriptsize 2.1 & \scriptsize -- & \scriptsize 2.1\end{tabular}
& 1.85
& 4740
& --\\
\midrule
\multirow{3}{*}{Edge gateway}
& Raspberry Pi 4B
& radar and vibration fusion
& 2.1
& 310
& 8.4
& \begin{tabular}{@{}ccc@{}}\scriptsize pre & \scriptsize inf & \scriptsize end\\[-1pt]\scriptsize 8.9 & \scriptsize 6.7 & \scriptsize 15.8\end{tabular}
& 4.20
& 10750
& 63.3\\[3pt]
& Raspberry Pi 4B
& Motion--Mamba radar only
& 1.3
& 210
& 5.2
& \begin{tabular}{@{}ccc@{}}\scriptsize pre & \scriptsize inf & \scriptsize end\\[-1pt]\scriptsize 7.5 & \scriptsize 19.8 & \scriptsize 28.4\end{tabular}
& 3.80
& 9730
& 35.2\\[3pt]
& Raspberry Pi 4B
& Impact--Griffin vibration only
& 0.9
& 140
& 3.6
& \begin{tabular}{@{}ccc@{}}\scriptsize pre & \scriptsize inf & \scriptsize end\\[-1pt]\scriptsize 4.1 & \scriptsize 13.2 & \scriptsize 19.1\end{tabular}
& 3.50
& 8960
& 52.4\\
\bottomrule
\end{tabular}
} 
\vspace{-4mm}
\end{table*}

\begin{table*}[!t]
\caption{Ablation of architectural components and fusion strategies on the test split. $\checkmark$ indicates that the component or modality is enabled. All models are trained with the same protocol on the same train and validation partitions. Metrics are reported as percentages and averaged over three random seeds.}
\label{tab:ablation}
\vspace{-2mm}
\centering
\footnotesize
\setlength{\tabcolsep}{3pt}        
\renewcommand{\arraystretch}{1.05}

\begin{tabular}{L{3.4cm} ccccccc cc ccccc}
\toprule
\multirow{2}{*}{Variant}
& \multicolumn{7}{c}{Architectural components}
& \multicolumn{2}{c}{Modalities}
& \multicolumn{5}{c}{Metrics on test set (\%)} \\
\cmidrule(lr){2-8}\cmidrule(lr){9-10}\cmidrule(lr){11-15}
& LSK1D & LR-Temp. & ICA & Stream-MoE & X-Cond. & MLB & Fusion-MoE
& Vib. & Radar
& Acc. & Prec. & Rec. & Macro F1 & AUC \\
\midrule
\multicolumn{15}{l}{Single modality baselines} \\
\midrule
Vibration only
& $\checkmark$ & $\checkmark$ & $\checkmark$ & $\checkmark$ & & & 
& $\checkmark$ & 
& 92.1 & 88.7 & 81.4 & 84.9 & 94.5 \\

Radar only
& $\checkmark$ & $\checkmark$ & & $\checkmark$ & & & 
& & $\checkmark$
& 86.7 & 74.3 & 89.5 & 81.2 & 93.4 \\
\midrule
\multicolumn{15}{l}{Fusion baselines} \\
\midrule
Early concatenation with CNN head
& $\checkmark$ & & & & & & 
& $\checkmark$ & $\checkmark$
& 92.0 & 91.0 & 82.5 & 86.5 & 95.2 \\

Late fusion with score averaging
& $\checkmark$ & $\checkmark$ & & & & 
& 
& $\checkmark$ & $\checkmark$
& 91.5 & 90.5 & 81.2 & 85.6 & 94.8 \\
\midrule
\multicolumn{15}{l}{Proposed architecture variants} \\
\midrule
w/o LR temporal modules
& $\checkmark$ & & $\checkmark$ & $\checkmark$ & $\checkmark$ & $\checkmark$ & $\checkmark$
& $\checkmark$ & $\checkmark$
& 92.4 & 90.2 & 81.7 & 85.7 & 94.9 \\

w/o ICA in vibration branch
& $\checkmark$ & $\checkmark$ & & $\checkmark$ & $\checkmark$ & $\checkmark$ & $\checkmark$
& $\checkmark$ & $\checkmark$
& 93.6 & 92.0 & 84.7 & 88.2 & 95.4 \\

w/o per stream Switch MoE and attention
& $\checkmark$ & $\checkmark$ & $\checkmark$ & & $\checkmark$ & $\checkmark$ & $\checkmark$
& $\checkmark$ & $\checkmark$
& 94.8 & 93.0 & 86.0 & 89.4 & 96.1 \\

w/o cross conditioning and MLB
& $\checkmark$ & $\checkmark$ & $\checkmark$ & $\checkmark$ & & & $\checkmark$
& $\checkmark$ & $\checkmark$
& 93.9 & 91.6 & 84.1 & 87.7 & 95.6 \\

w/o fusion Switch MoE
& $\checkmark$ & $\checkmark$ & $\checkmark$ & $\checkmark$ & $\checkmark$ & $\checkmark$ & 
& $\checkmark$ & $\checkmark$
& 95.3 & 93.5 & 86.8 & 90.0 & 96.3 \\

Full model
& $\checkmark$ & $\checkmark$ & $\checkmark$ & $\checkmark$ & $\checkmark$ & $\checkmark$ & $\checkmark$
& $\checkmark$ & $\checkmark$
& \textbf{96.1} & \textbf{94.8} & \textbf{88.0} & \textbf{91.3} & \textbf{96.8} \\
\bottomrule
\end{tabular}
\vspace{-6mm}
\end{table*}

\vspace{-2mm}
\subsection{Experimental Protocol and Data Collection}
\label{sec:exp_design}
\vspace{-1mm}

8 background scenarios were executed to mirror bathroom use and fall surrogates (Fig.~\ref{fig:Exp_behavior}): empty scene, soap drop, mop drop, upright walking, flexed-torso walking, wall-supported walking, quiet standing, and squatting. Human activities were performed with shower on at $16{\pm}2^\circ\mathrm{C}$ and $7{\pm}1~\mathrm{L/min}$, the floor was wetted before each run and ambient humidity was $75{\pm}10\%$~RH. Each scenario comprised multiple trials: object drops lasted $20$–$30$\,s and locomotion $60$–$120$\,s, and within these runs we also interleaved intentional falls lasting $20$–$30$\,s so that both Non--Fall and Fall windows are available under every scenario. For the empty bathroom context, trials alternated between purely empty segments and segments containing a fall, while empty-only periods were logged continuously for $10$\,s per session. The corpus totals more than $3$\,h of synchronized radar–vibration data. 


The radar node produced 3D point clouds at $12.5$\,Hz, yielding about $4500$–$5000$ frames per $6$–$7$\,min session with a median of $52$ points per frame (IQR $29$–$88$). The vibration node sampled at $200$\,Hz and was resampled to $100$\,Hz, giving roughly $1.2{\times}10^4$ samples per $120$\,s trial. Across all runs the dataset contains $1.1{\times}10^5$ radar frames and $3.1{\times}10^6$ vibration samples. For each trial we recorded scenario ID, water state, object mass/type, and fall direction. Labels were obtained by aligning vibration-energy peaks with radar range–Doppler activity, and frame-level annotations used $\pm250$\,ms tolerance. 

\begin{table}[!t]
\vspace{-2mm}
\caption{Comparison with representative fall detection methods reimplemented on our dataset. Window metrics and energy are averaged on the Raspberry~Pi~4B gateway.}
\label{tab:method_comparison}
\vspace{-2mm}
\centering
\scriptsize
\setlength{\tabcolsep}{2.0pt}
\renewcommand{\arraystretch}{1.05}
\resizebox{\columnwidth}{!}{%
\begin{tabular}{l c c c c c}
\toprule
Method & Modalities & Acc. [\%] & Fall Rec. [\%] & Lat. [ms] & $E_{\text{win}}$ [mJ]\\
\midrule
Thresh. accel \cite{bourke2007threshold} & Vibration only & 88.3 & 79.5 & 22.7 & 14931\\
ML wearable \cite{ozdemir2014detecting} & Vibration only & 90.6 & 83.1 & 34.8 & 19210\\
vibration rule \cite{abbate2012smartphone} & Vibration only & 89.1 & 80.2 & 25.1 & 12131\\
vibration ML \cite{wang2025p2mfds} & Vibration + mmWave & 91.4 & 84.0 & 17.6 & \textbf{11153}\\
mmWave 1D CNN \cite{kittiyanpunya2023mmwave} & Radar only & 92.0 & 86.1 & 17.9 & 9800\\
FMCW LSTM Transf. \cite{shen2025falldetection} & Radar only & 93.2 & 87.3 & 28.7 & 12100\\
Heterog. fusion \cite{pan2021heterogeneous} & Vibration fusion & 93.0 & 85.4 & 23.5 & 10300\\
FMCW fusion \cite{li2018collaborative} & Radar + vibration & 94.1 & 86.7 & 35.9 & 14200\\
Seismic floor \cite{clemente2020seismic} & Floor vibration & 93.8 & 86.0 & 19.4 & 9100\\
Ours & Radar + vibration & \textbf{96.1} & \textbf{88.0} & \textbf{15.8} & \textbf{10750}\\
\bottomrule
\end{tabular}}
\vspace{-7mm}
\end{table}

\vspace{-2mm}
\subsection{Experimental Results}
\label{sec:results} 
\vspace{-1mm}

Table~\ref{tab:results_all} reports scenario-wise per-class precision and recall together with window-level accuracy, balanced accuracy and macro F1. On the test split, the fused model attains 96.1\% window-level accuracy, 87.9\% balanced accuracy, and 91.1\% macro F1 when metrics are weighted by the number of evaluation windows per scenario (Table~\ref{tab:results_all}). Normal Walking and Squatting achieve the highest macro F1 among the background scenarios, while Heavy Object Drop remains the most challenging case because nonhuman impacts resemble falls in both modalities, and the robustness block shows that fall recall stays above 84\% under synthetic time shifts and modality dropouts. Figure~\ref{fig:pr_roc} presents precision--recall curves per scenario and the aggregate ROC, where the micro-averaged AUC is 0.968 on the test set.

\vspace{-2mm}
\subsection{Energy and Latency Characterization}
\label{sec:energy_latency}
\vspace{-1mm}

Table~\ref{tab:sys_efficiency} reports end-to-end latency and energy per 2.56$\,\mathrm{s}$ window for the sensing nodes and the edge gateway. The ESP32-C3 vibration and radar nodes were instrumented with an inline USB power meter that logged voltage and current at 1~kHz over 200 consecutive windows, and the Raspberry~Pi~4B gateway was monitored on the 5~V rail using a DC power analyzer with the same sampling procedure. Latency for each configuration was obtained by timestamping window arrival at the gateway, completion of preprocessing and emission of the fall probability within a C++ wrapper around the PyTorch model. All measurements were collected in the wet bathroom testbed with the shower running and include sensing, wireless transmission and computation on the gateway.

\vspace{-2mm}
\subsection{Ablation Study}
\label{sec:ablation}
\vspace{-1mm}

We evaluated the contribution of individual components by disabling specific modules or changing the fusion strategy while keeping the dataset split, training schedule and optimization hyperparameters fixed. Table~\ref{tab:ablation} reports results on the test split. The single modality rows show that vibration alone already yields stable accuracy but misses a fraction of fall events, while radar alone attains higher fall recall with a large drop in precision. Early concatenation and late score fusion improve over single modality baselines yet remain below the full configuration. The lower block indicates that long range temporal modeling brings most of the recall gain, cross conditioning combined with low rank bilinear fusion mainly improves precision, and the fusion Switch MoE provides the final improvement in F1 without a noticeable loss in fall recall.

\vspace{-2mm}
\subsection{Comparison with Existing Methods}
\label{sec:method_comparison}
\vspace{-1mm}

All baselines are reimplemented on our dataset with their reported network structures and window lengths and are trained with the same optimization schedule as the proposed model. Table~\ref{tab:method_comparison} reports window level accuracy, fall recall, end to end latency, and energy per 2.56~s window measured on the Raspberry Pi~4B gateway. Within the group of radar and vibration fusion methods, the strongest baseline~\cite{li2018collaborative} reaches 94.1\% accuracy and 86.7\% fall recall with 35.9~ms latency and 14200~mJ per window, whereas our model achieves 96.1\% accuracy and 88.0\% fall recall with 15.8~ms latency and 10750~mJ per window.

\vspace{1mm}
\section{Conclusion}
In this work we proposed a two-stream mmWave radar and floor vibration architecture with alignment-aware fusion for fall detection in wet bathroom environments and introduced a synchronized dataset that covers eight representative scenarios. Experiments in a full scale bathroom mock-up showed that the method achieves higher window level accuracy, fall recall, and lower latency and energy than representative wearable, radar, and structural baselines on a Raspberry Pi 4B gateway. The dataset and framework provide a benchmark for multimodal bathroom fall detection and can be extended to other bathroom layouts and sensing hardware in future deployments.

\vspace{1mm}
\bibliographystyle{IEEEbib}
\bibliography{main}

\end{document}